%% file: icme2023template.tex
\documentclass{article}
\usepackage{spconf,amsmath,epsfig}
\usepackage[noabbrev]{cleveref}
\usepackage{booktabs}
\usepackage{xcolor}
\usepackage{url}
\usepackage{amsfonts}  

\usepackage{lipsum}

\newcommand\CODE{\small\url{https://github.com/072jiajia/efficient-fld}}

\let\OLDthebibliography\thebibliography
\renewcommand\thebibliography[1]{
  \OLDthebibliography{#1}
  \setlength{\parskip}{0pt}
  \setlength{\itemsep}{0pt plus 0.3ex}
}

\begin{document}\sloppy

\def\x{{\mathbf x}}
\def\L{{\cal L}}

\title{Efficient Facial Landmark Detection for Embedded Systems}
%

\name{Ji-Jia Wu}
\address{jijiawu.cs@gmail.com}


\maketitle

\input{sections/0_abstract}

\input{sections/1_intro}

\input{sections/2_related_work}

\input{sections/3_method}

\input{sections/4_exp}
\input{sections/5_conclusion}

{
    \small
    \bibliographystyle{splncs04}
    \bibliography{icme2023template}
}

\end{document}

%% file: sections/0_abstract.tex
\begin{abstract}

This paper introduces the Efficient Facial Landmark Detection (EFLD) model, specifically designed for edge devices confronted with the challenges related to power consumption and time latency.
EFLD features a lightweight backbone and a flexible detection head, each significantly enhancing operational efficiency on resource-constrained devices.
To improve the model's robustness, we propose a cross-format training strategy. This strategy leverages a wide variety of publicly accessible datasets to enhance the model's generalizability and robustness, without increasing inference costs. Our ablation study highlights the significant impact of each component on reducing computational demands, model size, and improving accuracy.
EFLD demonstrates superior performance compared to competitors in the IEEE ICME 2024 Grand Challenges PAIR Competition, a contest focused on low-power, efficient, and accurate facial-landmark detection for embedded systems, showcasing its effectiveness in real-world facial landmark detection tasks.
The code is available at \CODE.
\end{abstract}

%% file: sections/1_intro.tex
\vspace{-0.25cm}
\section{Introduction}
\label{sec:intro}
Facial landmark detection plays a crucial role in the face processing pipeline, serving key applications such as virtual face reenactment, emotion recognition, and driver status monitoring. Despite the advancements and successes of deep learning models in addressing these challenges, deploying them on edge devices, especially smartphones, presents significant issues. These challenges, including power consumption and time latency, can severely degrade the user experience.

Existing works \cite{bulat2017far, kumar2020luvli, wang2020deep, zhu2021improving} propose methods that first predict a heatmap for each facial landmark, followed by post-processing each heatmap to determine the corresponding facial landmark position. These methods encounter long inference times and high computational costs due to their dense prediction, especially for models using an hourglass structure \cite{newell2016hourglass}.
To reduce the model's resource requirements, the Practical Facial Landmark Detector (PFLD) \cite{guo2019pfld} was introduced. It directly predicts the position of each facial landmark, thereby reducing the computational cost of dense predictions and eliminating the need for post-processing. While compromising on prediction precision, PFLD achieves a promising balance between prediction accuracy and computational cost, making it suitable for edge devices. However, further reduction in resource requirements remains a significant challenge.

In addition to computational demands on edge devices, facial landmark detection faces the fundamental challenge of model robustness. A straightforward approach to enhance robustness is to augment the training data with additional publicly available datasets, allowing the model to learn from a more diverse range of data. Nevertheless, this approach is not directly applicable in this task due to variations in landmark formats across different datasets. For instance, the WFLW dataset \cite{wu2018wflw} employs a 98-point format, Menpo \cite{zafeiriou2017menpo} uses a 68-point format, and the IVSLab Facial-Landmark Detection Dataset (IVSFD), provided by the competition host, adopts a 51-point format. While these datasets offer valuable training data, the discrepancies in landmark formats present a challenge in leveraging data from multiple sources.

To address these challenges, we propose the Efficient Facial Landmark Detection (EFLD) model, which consists of three key components: the efficient backbone network, the flexible facial landmark detection head network, and the cross-format training strategy. The efficient backbone and the flexible facial landmark detection head are designed to minimize time latency, power consumption, and memory usage on edge devices. The cross-format training strategy, on the other hand, is proposed to efficiently utilize existing publicly available datasets, thereby enhancing the generalizability of our model without increasing the inference cost.

The main contributions of this work are as follows:
\begin{itemize}
\item We introduce an efficient backbone network structure and a facial landmark detection head network structure that reduce time latency, power consumption, and memory usage on edge devices.
\item We propose a novel cross-format training strategy that utilizes existing publicly available datasets, even with different facial landmark formats, to enhance model performance and generalizability without increasing the inference cost.
\item EFLD demonstrates superior performance in the IEEE ICME 2024 Grand Challenges PAIR Competition, a contest focuses on Low-power, Efficient, and Accurate Facial-Landmark Detection for Embedded Systems, showcasing its effectiveness in real-world facial landmark detection tasks.
\end{itemize}

%% file: sections/2_related_work.tex
\begin{figure*}[t]
    \centering
    \includegraphics[width=0.95\linewidth]{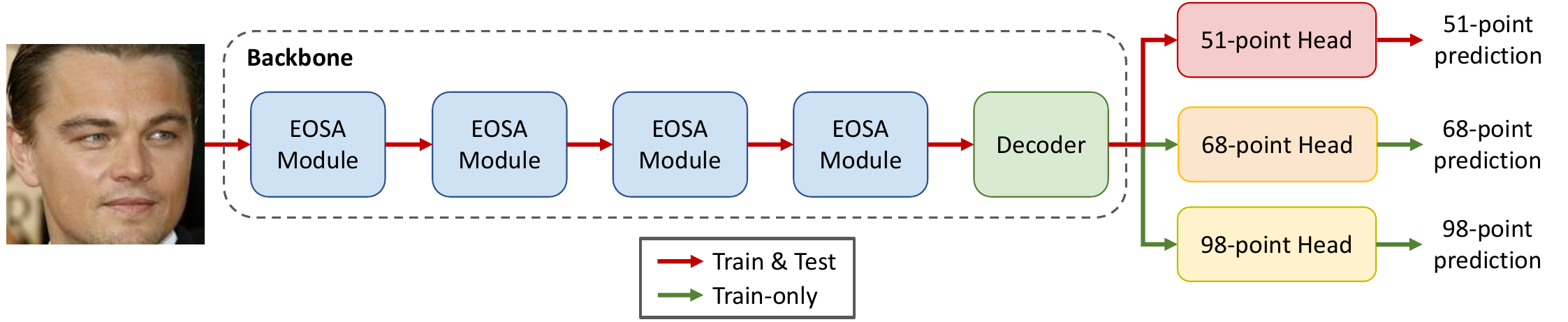}
    \vspace{-0.2cm}
    \caption{\small 
    \textbf{Model overview of our proposed method.} 
    Our method consists of three major components: (a) an efficient backbone network that transforms each input image into a feature vector, (b) multiple facial landmark detection head networks predicting facial landmarks in various formats, and (c) a cross-format training strategy that supports training across different facial landmark formats. When exporting our model for application, we only include the 51-point component in the exported model.
    }
    \label{fig:overallpipeline}
\end{figure*}

\section{Related Works}
\label{sec:related_works}

\subsection{Facial Landmark Detection}

Facial landmark detection has significantly evolved \cite{khabarlak2021fldsurvey} over the years, giving rise to the development of two primary methodological approaches: direct landmark regression and heatmap-based methods.
Direct landmark regression methods \cite{guo2019pfld} predict the positions of facial landmarks directly from images, providing faster inference times and suiting them for edge devices. Conversely, heatmap-based methods \cite{bulat2017far, kumar2020luvli, wang2020deep, zhu2021improving} produce 2D heatmaps for each landmark, subsequently post-processed to determine the corresponding facial landmark position. Although this approach can result in more accurate predictions, it demands more computational resources due to its dense predictions and the additional processing steps involved.

For our proposed method, we adopt direct landmark regression, favored for its rapid inference time. This choice aligns with our objective to enhance operational efficiency while maintaining high accuracy in landmark detection.

\subsection{Lightweight Model Architecture}

Deep learning models have demonstrated remarkable capabilities across a diverse range of tasks. However, deploying these models on edge devices presents significant challenges, particularly in terms of power consumption and time latency, which are critical constraints for devices like smartphones.

To mitigate these issues, researchers have developed several innovative strategies in recent years, including model quantization, efficient operations, network pruning, knowledge distillation, and neural architecture search, etc.
In our approach, we utilize the following two methods: \\
\textbf{Model Quantization:} This technique involves reducing the numerical precision of the model's parameters from floating-point to lower-bit representations, such as 16-bit half-precision or 8-bit integers. Despite a potential drop in performance due to the reduced precision, this modification significantly lowers the computational demands and memory usage of the model. \\
\textbf{Efficient Operations:} We reduce the overall computational complexity of our models by employing techniques such as depthwise separable convolutions \cite{chollet2017depthwiseseparable} and feature concatenation \cite{huang2017densenet, lee2019vov}. Depthwise separable convolutions reduce computational cost by dividing a standard convolutional layer into a combination of a 1-by-1 convolution and a depthwise convolution. For feature concatenation, we specifically employ the One-Shot Aggregation (OSA) Module \cite{lee2019vov}, which concatenates each layer's output to reduce computational cost. These methods significantly enhance the efficiency of our model.

%% file: sections/3_method.tex
\begin{figure*}[t]
    \centering
    \includegraphics[width=0.95\linewidth]{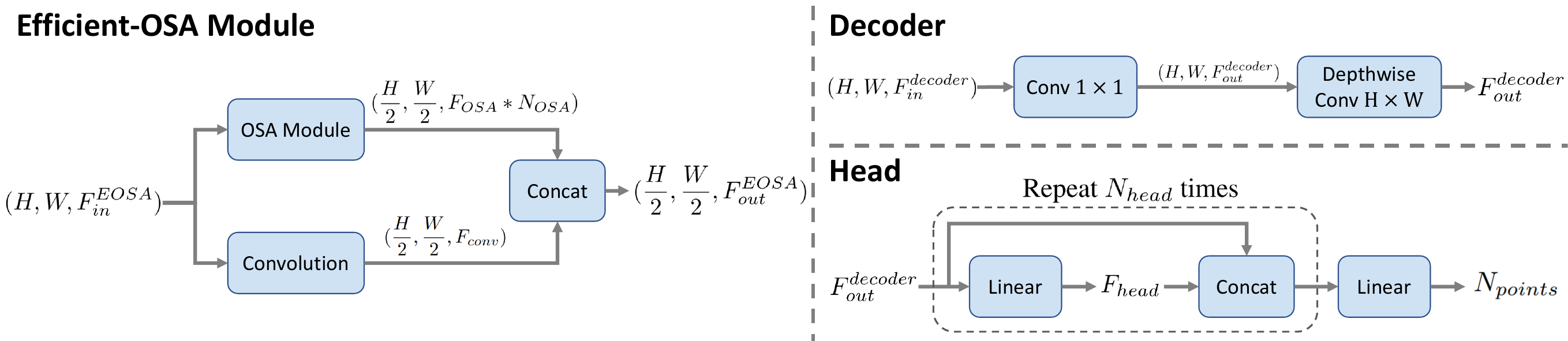}
    \caption{\small 
    \textbf{Detailed architecture of each components}
    The detailed architecture of the proposed Efficient-OSA (EOSA) module, decoder, and the facial landmark detection head network. Each module is lightweight due to the use of efficient operations such as feature concatenation layers and depthwise convolutional layers.
    }
    \label{fig:components}
\end{figure*}

\section{Methodology}
\label{sec:methodology}

In this section, we first explain the comprehensive methodology underpinning our proposed model in  \cref{subsec:overview}. Then, we specify the three major components of our method, including 1) the efficient backbone network structure in \cref{subsec:backbone}, 2) the flexible facial landmark detection head network in \cref{subsec:head}, and 3) the cross-format training strategy in \cref{subsec:strategy}. These components work harmoniously to ensure the model's performance is both lightweight for edge devices and robust for the performance.

\subsection{Method Overview}
\label{subsec:overview}
As illustrated in \cref{fig:overallpipeline}, our proposed method employs the direct landmark regression approach. Initially, the input image is processed through the backbone model, which consists of four Efficient-OSA (EOSA) modules, a modified-version of OSA module. The resultant feature map from the backbone is then decoded into a feature vector that represents the facial features. This feature vector is subsequently input into various facial landmark detection head networks, each tailored to generate predictions for different landmark formats. We apply the cross-format training strategy, optimizing our model on multiple publicly available datasets with diverse landmark formats simultaneously. When deploying our model, we only export the modules necessary for the application and remove unrelated modules, thus reducing the required resources.

\subsection{The Lightweight Backbone Network}
\label{subsec:backbone}
Our backbone network consists of four EOSA modules followed by a decoder. As illustrated in \cref{fig:components}, each EOSA module inputs a feature map and outputs a downscaled feature map with stride of 2. This structure incorporates elements of OSA module \cite{lee2019vov} and adds an extra convolutional layer.
The design aims to harness OSA module's lightweight and efficient computation capabilities to extract more complex features, while the added convolutional layer focuses on capturing low-level features.
To obtain the downscaled feature map, the first convolutional layer of the OSA module has a stride of 2, followed by $N_{blocks} - 1$ convolutional layers with a stride of 1. Each layer outputs a downscaled feature map with a channel number of $F_{OSA}$.
Furthermore, for the added convolutional layer, we employ depthwise separable convolutional layers to further decrease the computational cost of OSA module. This integration effectively reduces the model size with only a minimal impact on performance.

The decoder within the backbone is designed straightforwardly. It comprises a $1 \times 1$ convolutional layer with an output channel number of $F^{decoder}_{out}$, followed by a depthwise convolution layer with a filter size of $H \times W$, where $H$ and $W$ represent the height and width of the feature map, respectively. Thus, the depthwise convolution layer outputs a feature vector with dimension $F^{decoder}_{out}$, representing the features of the face. This setup decodes the feature map into a feature vector through the combination of $1 \times 1$ convolution and depthwise convolution, thereby reducing computational resources.

\subsection{The Facial Landmark Detection Head}
\label{subsec:head}
As depicted in \cref{fig:components}, each detection head takes a feature vector representing the face and outputs the positions of facial landmarks. Inspired by \cite{huang2017densenet}, the first part of the architecture consists of $N_{head}$ linear layers followed by concatenation steps. Each linear layer outputs a feature vector with a dimension of $F_{head}$, which is then concatenated with the input feature vector. The second part of the architecture is a linear layer responsible for predicting the positions of all landmarks based on the concatenated feature vector. Notably, an EFLD model may include multiple detection head networks, each tailored for predicting a specific type of facial landmark. This design enables the model to predict multiple types of facial landmarks from a single input image, significantly enhancing the versatility and utility of the EFLD model across various applications. Additionally, when exporting our model to applications, since each detection head is responsible for a single type of facial landmark format, we may only export the necessary heads to our applications, thereby eliminating unnecessary computations.

\subsection{The Cross-Format Training Strategy}
\label{subsec:strategy}
The cross-format training strategy is designed to improve the model's generalizability and robustness by joint-training on multiple datasets.
Specifically, for training on multiple datasets simultaneously, we create a detection head for each type of facial landmark format present in the training data. These datasets are then shuffled, and batch training is applied to this training pipeline.
This approach addresses the diversity in facial landmark formats, such as the 98-point format used by the WFLW dataset \cite{wu2018wflw}, the 68-point format of Menpo \cite{zafeiriou2017menpo}, and the 51-point format adopted by the IVSLab Facial-Landmark Detection dataset (IVSFD), which is provided by the competition host. Our model undergoes simultaneous training on these datasets. With multiple detection heads tailored for specific landmark formats, including 51-point, 68-point, and 98-point, the model calculates the loss function for annotations when available and assigns a zero loss for missing annotations.

This strategy enables efficient and effective learning from a wide range of data sources, significantly enhancing the model's ability to perform consistently across different landmark configurations. Additionally, this strategy is highly flexible, as the additional training resources can be extended without limit. It allows the model to adapt and accurately predict facial landmarks in a wide array of facial expressions, environments, and landmark formats, making it highly effective in diverse real-world applications.

%% file: sections/4_exp.tex
\input{tables/sota}

\section{Experiments}
\label{sec:experimental_results}

\subsection{Implementation Details}
\label{subsec:details}
Our training pipeline is developed upon TF-Models \cite{tensorflowmodelgarden2020tfm}. \\
\textbf{Model Hyperparameters.} The configuration for the four EOSA modules in the backbone network, $(F_{OSA}, N_{OSA})$, is set as $(4, 2)$, $(8, 3)$, $(16, 3)$, $(16, 3)$, with only the first EOSA module utilizing conventional convolution with $F_{conv}=8$. The remaining modules employ depthwise separable convolutional layers with $F_{conv} = 8$, $F_{conv} = 16$, and $F_{conv} = 32$, respectively. The output feature dimension of the decoder is set to $256$. The detection head is composed of three linear blocks, each with $32$ units, followed by concatenation. \\
\textbf{Training datasets.} Our model is trained across multiple datasets: the IVSFD dataset, which includes 4437 images; the training set of Menpo \cite{zafeiriou2017menpo}, comprising 6655 images; and the WFLW dataset \cite{wu2018wflw}, totaling 10000 images. \\
\textbf{Training Hyperparameters.} We apply the Normalized Mean Error (NME), normalized by the inter-ocular distance, as our training loss. The input images are resized to 128, and trained with batch size of $512$ on one NVIDIA $2080$Ti GPU, with learning rate at $0.001$ using the AdamW \cite{loshchilov2017adamw} optimizer. The training spans $1500$ epochs, utilizing a cosine schedule for the learning rate adjustment. \\
\textbf{Export TFLite Model.} For the final version of our model, we export only the backbone model along with the 51-point detection head. To optimize our model's performance, we implement int8 post-training quantization, conducting calibration on all images from the IVSFD dataset. \\
\textbf{Inference.} For each input data, we first utilize MediaPipe \cite{lugaresi2019mediapipe}, which provides a cross-platform face detection function to identify faces in the images. Then, our TFLite model is used to predict the facial landmarks for each detected face.

\subsection{Performance of Final Submission}

Table 1. presents the results of our proposed method alongside the top ten competitors in the final round of the IEEE ICME 2024 Grand Challenges PAIR Competition. This table details each method's performance on the private testing set provided by the competition host, evaluated on the MediaTek Dimensity 9300 Series Platform. The reported metrics include accuracy, model complexity, size, inference time, power consumption, and the application of int8-quantization. 
The accuracy metric, designed by the competition host, is measured using a per-point evaluation criterion: a prediction is determined to be correct if the predicted facial landmark falls within 5 pixels of the ground truth; otherwise, it is determined to be incorrect.
Additional metrics include model complexity, measured in GFLOPs; model size, reported in megabytes; inference time per image, in milliseconds; and power consumption per image, in watts.
The final score for each method is calculated based on the formula provided by the host:
\begin{equation}
    Score = \frac{Accuracy}{Time \cdot Complexity \cdot Power \cdot Size}
\end{equation}

Our method achieved the highest final score, significantly outperforming the runner-up. We also recorded the best performance in terms of model complexity, size, inference time, and power consumption. Notably, our team was the only one to implement int8-quantization, which effectively reduced costs while still maintaining acceptable performance of accuracy.
These results demonstrate the effectiveness of our Efficient Facial Landmark Detection (EFLD) method in providing a lightweight and efficient solution for facial landmark detection on edge devices.

\input{tables/ablation_study}
\subsection{Ablation Study}
Table 2. showcases the impact of different components on the model's performance by comparing Normalized Mean Error (NME), Failure Rate at 10\% (FR@10\%), and Cumulative Error Distribution - Area Under Curve at 10\% (CED-AUC@10\%), with NME normalized by the inter-ocular distance.
In these experiments, we split the IVSFD dataset provided by the competition host into a training set with 2218 samples and an evaluation set with 2219 samples. Additionally, when employing the cross-training strategy, we incorporate samples from the Menpo \cite{zafeiriou2017menpo} and WFLW \cite{wu2018wflw} datasets.
In the table, ``L. Backbone'' signifies the adoption of depthwise separable convolution within the EOSA module of the backbone network, in contrast to using a standard convolutional layer. ``L. Head'' denotes the use of our proposed detection head as opposed to PFLD's detection head \cite{guo2019pfld}. ``Cross-Format'' refers to the utilization of additional publicly available datasets for training our model.

The results underscore the significant efficiency of the lightweight backbone through reduced Million FLOPs, indicating its crucial role in decreasing computational demands. The lightweight head not only minimizes model size but also sustains competitive performance, underscoring its effectiveness. Moreover, the cross-format training strategy significantly enhances model accuracy and performance, highlighting its essential contribution to improving model efficacy.

In essence, our ablation study validates the vital significance of each component in enhancing the model's performance, affirming their collective influence on computational efficiency, size reduction, and accuracy improvement.

%% file: tables/sota.tex
\begin{table*}[t]
\centering
\renewcommand\arraystretch{1.2} 
\renewcommand\tabcolsep{15pt} 
{%
\begin{tabular}{c|cccccc|c}
\toprule
Team & Accuracy & G FLOPs  &  Size &  Time & Power & INT8  & Score  \\
\midrule
jijiawu (Ours) & 18.47 & \textbf{0.02} & \textbf{0.17} & \textbf{1.08} & 1.93 & \checkmark & \textbf{2741.92} \\
\midrule
yuhs1 & 19.30 & 0.08 & 2.25 & 8.12 & \textbf{1.64} &  & 7.78\\
JordanHsieh & 19.82 & 0.15 & 1.58 & 6.98 & 1.79 &    & 6.86 \\
hiwu.dery & 23.10 & 0.15 & 1.46 & 42.97 & 2.25 &     & 1.10 \\
O\_S\_O   & 25.50 & 0.38 & 2.48 & 15.85 & 2.07 &        & 0.83\\
ppp458732 \cite{hong2024improving} & 22.30 & 1.16 & 2.64 & 67.35 & 2.27 &     & 0.048 \\
Weiiiii\_0123 & \textbf{28.06} & 11.15 & 4.12 & 17.55 & 2.04 &  & 0.017 \\
dar814   & 4.09 & 0.15 & 10.00 & 26.10 & 9.34 &     & 0.011  \\
YUJIE47  & 16.13 & 2.61 & 4.55 & 151.55 & 9.84 &      & 0.0008 \\
asdggg & 19.95 & 4.34 & 28.46 & 110.03 & 2.70 & & 0.0005 \\
ed1122 & 16.96 & 33.29 & 54.15 & 87.77 & 2.87 &  & 0.00004 \\
\bottomrule
\end{tabular}
\vspace{-0.3cm}
}
\caption{The proposed method is compared with ten competitors' methods in the final round of the contest across various metrics: accuracy, model complexity (G FLOPs), model size (Size), inference time (Time), power consumption (Power), the application of int8-quantization (INT8), and final score (Score). The best method for each metric is indicated in bold.
}
\label{tab:sota}
\end{table*}

%% file: tables/ablation_study.tex
\begin{table*}[t]
\centering
\renewcommand\arraystretch{1.2} 
\renewcommand\tabcolsep{6pt} 
{%
\begin{tabular}{ccc|ccccccc}
\toprule
L. Backbone & L. Head & Cross-Format & M FLOPs  & Size & NME & FR@10\% & CED-AUC@10\%  \\
\midrule
\checkmark & \checkmark & \checkmark & 19.1 & 0.170 &
3.72 & 1.13 & 65.3 \\
\midrule
           & \checkmark & \checkmark & \textcolor{red}{25.2} & 0.180 &
3.68 & 1.26 & 65.9 \\
\checkmark &            & \checkmark & 18.8 & \textcolor{red}{0.229} &\
3.62 & 1.08 & 66.0 \\
\checkmark & \checkmark &            & 19.1 & 0.170 &
\textcolor{red}{5.10} & \textcolor{red}{4.42} & \textcolor{red}{54.2} \\
\bottomrule
\end{tabular}
}
\caption{\textbf{Ablation study.}
Performance of different combinations of proposed components, including the lightweight backbone (L. Backbone), lightweight facial landmark detection head (L. Head), and the application of the cross-format training strategy (Cross-Format). We report results across various metrics: model complexity (M FLOPs), model size (Size), normalized mean error (NME), failure rate at a threshold of 10\% (FR@10\%), and cumulative error distribution - area under curve at a threshold of 10\% (CED-AUC@10\%). The method with the largest performance drop for each metric is indicated in red.
}
\label{tab:ablation}
\end{table*}

%% file: sections/5_conclusion.tex
\section{Conclusions}
\label{sec:conclusions}

Our work advances the field of facial landmark detection by introducing the Efficient Facial Landmark Detection (EFLD) model, tailored for optimal performance on edge devices. By innovating an efficient backbone and a flexible detection head, we address the critical issues of high power consumption and latency that hinder deployment on such devices. Moreover, our novel cross-format training strategy successfully tackles the challenge of varying landmark formats across datasets, enhancing our model's generalizability without increasing the inference cost. Demonstrating superior performance against competitors, EFLD represents a significant step forward, promising improved user experience in real-world applications.